# Forecasting Sleep Apnea with Dynamic Network Models

**Paul Dagum**
Section on Medical Informatics
Stanford University School of Medicine
and
Rockwell Palo Alto Laboratory
444 High Street
Palo Alto, California 94301

**Adam Galper**
Section on Medical Informatics
Stanford University School of Medicine
Stanford, California 94305-5479

## Abstract

*Dynamic network models (DNMs)* are belief networks for temporal reasoning. The DNM methodology combines techniques from time-series analysis and probabilistic reasoning to provide (1) a knowledge representation that integrates noncontemporaneous and contemporaneous dependencies and (2) methods for iteratively refining these dependencies in response to the effects of exogenous influences. We use belief-network inference algorithms to perform *forecasting*, *control*, and *discrete-event simulation* on DNMs. The belief-network formulation allows us to move beyond the traditional assumptions of linearity in the relationships among time-dependent variables and of normality in their probability distributions. We demonstrate the DNM methodology on an important forecasting problem in medicine. We conclude with a discussion of how the methodology addresses several limitations found in traditional time-series analyses.

## 1 INTRODUCTION

Most probabilistic reasoning research has focused on the construction and use of fundamentally static models, in which custom-tailored temporal relationships are fixed for all times. The predictions of a static model are *time invariant* in that the model makes the same inferences regardless of when observations are made. The model is *memoryless* in that it uses only the set of current observations to predict the state of the system.

Although special-case temporal problems can be solved by such static models, in general, a true Bayesian consideration of a complex process unfolding in time must provide a means for representing and updating time-dependent probabilistic relationships. Compared to static models, a truly temporal model is enriched through a consideration of trend information, and has inferential machinery to update a model in response to changing information about the history of a process. We refer to the adaptation of a model to an ongoing process as *dynamic modeling*. Failure to adapt to relevant and emergent trends can severely bias a model.

We have pursued the development of dynamic models for temporal probabilistic reasoning. Such models can be applied to forecasting, control, and discrete-event simulation problems. In this paper, we focus on the task of forecasting, borrowing heavily from statistical forecasting techniques.

Statisticians have developed numerous methods for reasoning about temporal relationships among variables; the field is generally known as *time-series analysis*. A *time series* is a sample realization of a stochastic process, consisting of a set of observations made sequentially over time. Investigators have sought to develop time-series methods for generating or inducing stochastic models, which describe temporal dependencies among successive observations in a time series.

Time-series analyses have been quite successful; a mature set of Bayesian time-series analysis methods has been developed and applied to a wide range of problems [15, 31]. However, classical time-series methodologies are restricted in their ability to represent the general probabilistic dependencies and the nonlinearities of real-world processes. Investigators are forced to wrestle complex problems into relatively simple parametrized models that can be solved with the traditional methods. Until recently, there has been relatively little interaction between the statisticians interested in time-series analysis and the computer scientists studying the representation of uncertain knowledge with Bayesian belief networks.

In this paper, we describe a robust and expressive forecasting procedure based on the integration of fundamental methods of Bayesian time-series analysis with belief-network representation and inference techniques. This synthesis, embodied in the *dynamic network model* (DNM), has two immediate benefits. First, by casting Bayesian time-series analyses as temporal belief-network problems, we can introduce arbitrary dependency models that capture richer, and more realistic, models of dynamic dependencies—as well as the more traditional static (or *contemporaneous*) belief-network dependencies. A robust knowledge representation can simplify construction of dynamic models by making explicit many of the assumptions in the underlying modeling technique. Second, we can apply belief-network inference algorithms to the models to generate normative forecasts. The inference algorithms exploit the temporal representation, rendering inference tractable for large belief networks[8]. The richer models and associated computational methods allow us to move beyond such rigid classical assumptions as linearity in the relationships among variables and normality of their probability distributions.

We describe an implementation called **DYNEMO** and validate the DNM forecasting methodology through the analysis of a multivariate time series of 34,000 recordings of *sleep apnea* data[1]. Sleep apnea is a serious medical condition, characterized by intermittent periods of arrested breathing during asleep. The National Commission on Sleep Disorders Research, established by Congress in 1988, reported recently that each year the lives of millions of Americans are disturbed, disrupted, or destroyed by the consequences of sleep disorders. The most serious sleep disorder in terms of morbidity and mortality is obstructive sleep apnea [24]. Forecasts of apneic episodes could reduce potentially the morbidity and mortality associated with sleep apnea.

## 2 RELATED WORK

Although most temporal-reasoning research is based on logic [27], several frameworks have been proposed to support temporal reasoning using probability theory. Dean and Wellman [14] provide a good summary of probabilistic models for temporal reasoning and the modeling of dynamic domains. We summarize here some recent results that are relevant to the work presented.

Berzuini [4] embeds semi-Markov models in a belief-network representation and uses approximate probabilistic inference to compute the degree of belief in past states and in future states. The inability of these models to adapt to new observations, and their Markov nature, renders them unable to make forecasts that extend beyond Markov simulation. Dean and Kanazawa [12, 13] develop a probabilistic model for projection based on a functional (e.g., exponential) decay model of the persistence with time of propositions. Kanazawa [20] achieves a synthesis of Bayesian belief networks and elements of the temporal logics of Shoham [26], Bacchus [3], and Halpern [16]. Abramson [2] con-structs a belief-network model for forecasting crude-oil prices where the prior and conditional probabilities are obtained using external regression models. Kjaerulff [23] considers temporal belief networks that consist of a finite number of time slices connected by Markovian dependencies; he describes a more efficient method of junction-tree formation for these networks, but does not address how the model structure and conditional probabilities evolve over time. Approaches that do address dynamic modeling with Markovian probabilistic networks include those of Kenley [22] and Tatman and Shachter [28, 29].

## 3 THE DYNAMIC NETWORK MODEL

Previously we have developed a belief-network–based time-series model called the *dynamic network model (DNM)* [9]. In this section, we briefly summarize the structure of DNMs, and describe how we employ *convex combination* to specify conditional probabilities.

### 3.1 STRUCTURE

DNMs consist of nodes which represent domain variables at different time points—for example, heart rate at time $t$, $HR_t$ and heart rate at time $t-1$, $HR_{t-1}$ in Figure 1. Nodes are connected by *contemporaneous dependencies* if they represent variables at the same time point. Otherwise, nodes are connected by *noncontemporaneous dependencies*.

The DNM conditional probabilities are obtained through the convex combination of the contemporaneous and noncontemporaneous relations. For node $X_{it}$, let $\pi(X_{it})$ and $\theta(X_{it})$ denote the sets of contemporaneous and noncontemporaneous parents of $X_{it}$ in the DNM, respectively. The convex combination for contemporaneous and noncontemporaneous relations is

$$\begin{aligned}\Pr(X_{it} \mid \pi(X_{it}), \theta(X_{it})) &= (1-\alpha_{it}) \cdot \Pr(X_{it} \mid \pi(X_{it})) \\ &+ \alpha_{it} \cdot \Pr(X_{it} \mid \theta(X_{it})).\end{aligned} \quad (1)$$

The weight $\alpha_{it}$ in Equation 1 lies between 0 and 1, and expresses the relative confidence ascribed to the prior estimate of $X_{it}$ and the likelihood provided by the observations. When the weight is close to 1, the prior distribution is more informative than the likelihood provided by the observation; when the weight is close to 0, the prior distribution is less informative than the likelihood provided by the observation. Note that the weighting coefficients are time dependent, and the model adapts dynamically to changing exogenous influences by changing the weights.

The method of convex combination may appear initially to be a naive procedure for integrating historical information with current estimates of domain variables. The method, however, embodied in the form of the Kalman filter in state-space models, and in the conditional sum of squares in ARIMA models [17], is an integral aspect of models that purport to forecast future values of time series. In the absence of prior information, a convex combination is the simplest method of endowing the model with the capacity to adapt to changing exogenous influences.

---

[1]The data were collected from a patient in the sleep laboratory of the Beth Israel Hospital in Boston, Massachusetts, and made available by the Santa Fe Institute as part of a time-series competition held in the Fall of 1991 [30].

## 3.2 VALIDITY

In [9] we discuss log-linear decompositions, in addition to convex-combination, as a valuable decomposition of conditional probabilities. In general, we refer to these methods as *additive decompositions*. In a static model, an additive decomposition of conditional probabilities is an approximation of the true conditional probability. In a dynamic model, an additive parametrization is necessary for adaptability to unmodeled exogenous influences.

Additive decompositions overcome two difficult problems researchers encounter in large belief network applications [8]: intractable inference and intractable induction. The performance of exact inference deteriorates rapidly with increasing order because the non-contemporaneous dependencies yield very large clique sizes. However, Dagum and Galper [8] show that an exact inference algorithm can exploit the additive decomposition of the conditional probabilities to reduce the complexity of inference in a DNM. The complexity of inference is determined by the size of the largest clique contained in a single time slice, rather than the largest clique of the entire DNM. Similarly, automated induction of a large DNM with Cooper and Herskovits' algorithm [6] is possible if we adapt the algorithm to exploit the additive decomposition. We discuss this point further in Section 5.

## 3.3 MODEL UPDATE

In time-series analyses, dynamic models adapt to changes in system behavior through the reestimation of model parameters when new observations are made. The process of *updating* a DNM is the iterative process by which we use new observations to update the estimates of the weighting coefficients in Equation 1. In [10], we consider three methods of update: maximum likelihood, Bayesian update, and maximum expected utility. Maximum expected utility update produced the best forecast results in the analysis of the sleep-apnea data, and we discuss only this method here.

Maximum utility assumes access to a utility model, or equivalently, a loss function. We choose parameters that maximize the expected utility of the model, or equivalently, that minimize a loss function. In the analysis of the sleep-apnea data, we assume a quadratic loss function. Minimizing loss is similar to the non-Bayesian method of least squares estimation (LSE) of parameters. When the model has been properly fitted, the residuals $\epsilon_t$ between the forecasts and the observed values are mean zero, normally distributed, independent random variables, and as such, LSE of the parameters is equivalent to the determination of the parameters by maximizing the likelihood of the observed data [17].

To illustrate model update in DNMs, assume we want to estimate the parameter $\alpha$ at time $t$ to be used in forecasting the value of $Z_{t+1}$. If at time $t$, $Z_t$ was observed to have value $\hat{z}$, then the deviation $\epsilon_t$ is given by

$$\epsilon_t = \sum_{Z_t = z} z \Pr[Z_t = z | E_{t-1}] - \hat{z}.$$

Note that $\epsilon_t$ measures the deviation of the *expected* forecast value of $Z_t$ from the observed value $\hat{z}$. To estimate $\alpha$, LSE solves for the $\alpha$ that minimizes the sum

$$S(\alpha) = \sum_{i=0}^{t} \epsilon_i^2. \qquad (2)$$

In Equation 2, LSE weights equally all deviations of the forecast from the observed values. When we estimate the DNM parameters we use in a forecast, however, minimization of recent deviations is more critical than minimization of old deviations. To discount the contribution of historical deviations in the LSE of parameters we weight each $\epsilon_i^2$ in Equation 2 with a *geometric discount factor* $\theta^{t-i}$, where $\theta \leq 1$. This method, known as *discounted least squares (DLS)* [5, 1], is used extensively in model fitting of time-series models. For dynamic linear models, Brown [5] suggests values for $\theta$ between 0.7 to 0.95, slightly higher than what we found optimal for the sleep-apnea DNM.

## 4 THE DYNEMO IMPLEMENTATION

The Dynamic Network Modeler (**DYNEMO**) includes implementations of the K2 belief-network learning algorithm [6], temporal extensions to K2, lexical analyzers and parsers for processing various belief network and DNM file formats, exact and approximate probabilistic inference algorithms, and an assortment of algorithms for estimating DNM parameters. **DYNEMO** is written entirely in ANSI C and has been tested on several UNIX platforms. Graphical analysis of **DYNEMO** forecasts is performed in Mathematica [25].

The **DYNEMO** workbench provides an integrated, interactive environment for DNM generation and forecasting. The DNM generation module produces as output an external representation of a DNM. The DNM forecasting module takes as input the generated DNM.

### 4.1 DNM GENERATION

DNM generation requires as input the specification of model variables $X_1, X_2, \ldots, X_n$, including their cardinalities and discretizations, the desired DNM order $p$, and a historical database $D$, in which each case records the instantiations of model variables at some time $t$. Since DNMs and belief networks capture discrete probability distributions, all continuous variables must be discretized.

**DYNEMO** generates candidate DNM structures using a modified K2 belief-network learning metric [18]. The K2 metric scores a candidate network structure by searching for the parent set of each node that maximizes the likelihood of the observed data. The K2 algorithm accepts as input a database of records. Each record contains an instantiation of some set of nodes in the belief network. K2 assumes that the records are generated by an *independent* and *identically* distributed process. This assumption is valid when we consider static belief networks. However, in a temporal database, records are not independently distributed. Thus, in evaluating the K2 likelihood function, the joint probability of the records conditioned on the belief network structure decomposes into a product of likelihood functions. These likelihood functions express the probability of a single record conditioned on the belief network structure. When the records are *not* independently distributed, the K2 likelihood function decomposes into a product of likelihood functions for each record that are *conditioned* on a small set of historical observations, in addition to the belief network structure. The modified K2 algorithm maximizes the likelihood function for DNMs by employing the latter decomposition. In this way, the modified K2 algorithm captures the DNM structure that maximizes the likelihood of the data and their temporal crosscorrelation.

To compute the conditional probabilities, we transform the DNM structure into two belief networks, $BN_c$ and $BN_{nc}$, by selectively removing noncontemporaneous and contemporaneous dependencies, respectively. We then tally the cases in the database for each node in $BN_c$ and $BN_{nc}$, generating contemporaneous and noncontemporaneous conditional probabilities for the resultant DNM.

### 4.2 DNM FORECASTING

To output one- through $k$-step ahead forecast distributions of the model variables, **DYNEMO** takes as input a DNM, a database $D$, and a parameter update method.

Let $BN_{\alpha_t}^{t+1}$ denote the belief network generated by instantiating the parameters of the DNM to $\alpha_t$, the parameters estimated after observing evidence at time $t$.

For a one-step ahead forecast, **DYNEMO** generates forecasts by performing probabilistic inference on $BN_{\alpha_t}^{t+1}$. The posterior marginal distributions of the leading-slice nodes are the forecast distributions. For example, the network in Figure 1a depicts the topology of the sleep-apnea DNM and of the corresponding $BN_{\alpha_t}^{t+1}$.

For a $k$-step ahead forecast, **DYNEMO** recursively generates $BN_{\alpha_t}^{t+2}$, $BN_{\alpha_t}^{t+3}$,..., $BN_{\alpha_t}^{t+k}$. To generate $BN_{\alpha_t}^{t+i}$,

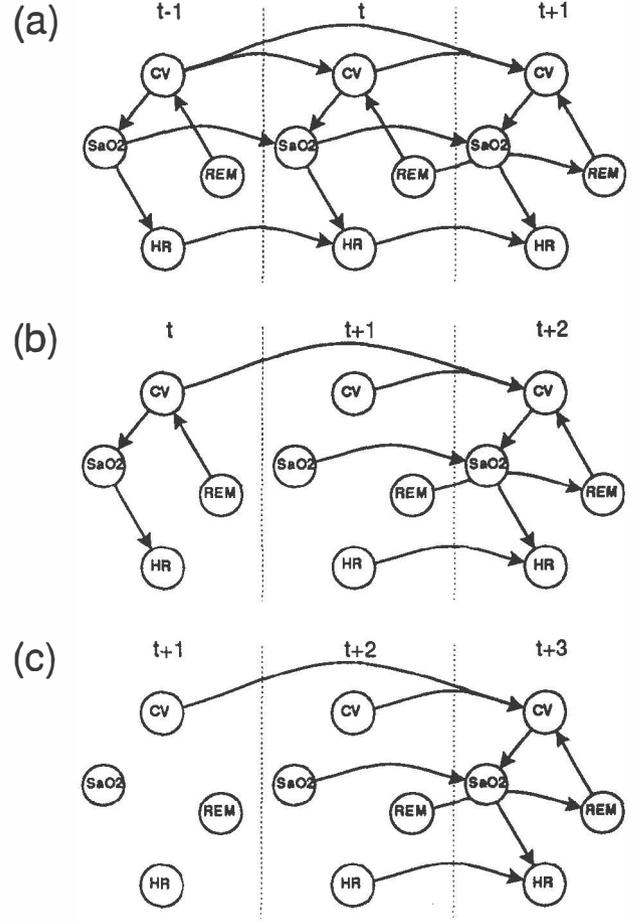

Figure 1: The sleep apnea DNM and its forecast belief networks. Each model contains nodes for heart rate (HR), chest volume (CV), oxygen saturation (SaO2), and sleep state (REM) over three time points. Straight arcs indicate contemporaneous dependencies; curved arcs are noncontemporaneous dependencies. (a) When **DYNEMO** instantiates this DNM with parameters $\alpha_t$, a topologically identical belief network $BN_{\alpha_t}^{t+1}$ results, from which one-step ahead forecasts can be generated. (b) The $BN_{\alpha_t}^{t+2}$ generated by removing arcs from the $BN_{\alpha_t}^{t+1}$ in (a). The leading-slice forecasts at time $t+1$ in $BN_{\alpha_t}^{t+1}$ are assigned as the prior distributions of $HR_{t+1}$, $CV_{t+1}$, $SaO2_{t+1}$, and $REM_{t+1}$ in $BN_{\alpha_t}^{t+2}$. This belief network is used to generate two-step ahead forecasts of the model variables. (c) The $BN_{\alpha_t}^{t+3}$ generated by removing arcs from the $BN_{\alpha_t}^{t+2}$ in (b). Here, the forecast distributions at time $t+1$ in $BN_{\alpha_t}^{t+2}$ become prior distributions at time $t+1$ in $BN_{\alpha_t}^{t+3}$. Likewise, the forecast distributions at time $t+2$ in $BN_{\alpha_t}^{t+2}$ become prior distributions at time $t+2$ in $BN_{\alpha_t}^{t+3}$. This belief network is used to generate k-step ahead forecasts ($k \geq 2$) of the model variables.

DYNEMO removes arcs from nodes in $BN_{\alpha_t}^{t+i-1}$ according to the following rule:

> If node $X_t$ is uninstantiated in $BN_{\alpha_t}^{t+i-1}$, then render node $X_{t-1}$ a prior node in $BN_{\alpha_t}^{t+i}$ with a prior distribution equal to the posterior marginal forecast of $X_t$ in $BN_{\alpha_t}^{t+i-1}$.

These structural changes may disconnect the network. For example, the DNM of Figure 1a generates the $BN_{\alpha_t}^{t+2}$ in Figure 1b; $BN_{\alpha_t}^{t+2}$ is then used to generate the disconnected network $BN_{\alpha_t}^{t+3}$ in Figure 1c. Higher-order forecasts are generated by performing probabilistic inference on the respective $BN_{\alpha_t}^{t+i}$, after instantiating the appropriate temporal evidence.

Let $l$ be the highest-order lag of a node in the leading time slice of a DNM. Note that there will be $\min(l, k)$ unique $BN_{\alpha_t}^{t+i}$ that must be constructed to generate $k$-step ahead forecasts.

DYNEMO provides both exact (Lauritzen-Spiegelhalter) and approximate (BN-RAS [7]) probabilistic inference for computing forecast distributions. Exact inference is more efficient than approximate inference when time series contain few missing values, in which case the $BN_{\alpha_t}^{t+i}$ are predominantly instantiated.

To update the DNM parameters, DYNEMO instantiates the leading-slice model variables in $BN_{\alpha_t}^{t+1}$ with freshly observed evidence and computes new parameter values using the update method specified by the user (see Section 3.3).

To visualize the results of forecasting, DYNEMO outputs expected values of the k-step ahead forecasts for graphical analysis within Mathematica[25]. Mathematica plots observed data versus forecast data and computes and plots the sum of the squares of the residuals (see Section 5).

## 5 THE SLEEP-APNEA FORECASTING PROBLEM

In 1991, the Santa Fe Institute organized a time-series forecasting competition [30]; one of the databases made available in that competition was a multivariate data set of 34,000 recordings, sampled at 2 Hz, of heart rate (HR), chest volume (CV), blood oxygen concentration (SaO2), and sleep state (REM). The data were collected from a patient suffering from sleep apnea in the sleep laboratory of the Beth Israel Hospital in Boston, Massachusetts.

### 5.1   SLEEP APNEA

Sleep apnea is a serious and prevalent medical condition, characterized by periods during which a patient takes a few quick breaths and then stops breathing for up to 45 seconds. This pattern is repeated as many as 200 to 400 times during six to eight hours of sleep. The main clinical consequence of sleep apnea is excessive daytime sleepiness related to the fragmentation of sleep and the effects of hypoxemia on cerebral function. Thus, for example, patients with severe obstructive sleep apnea are involved in automobile accidents two to three times more often than the general population. In addition to daytime sleepiness, studies implicate sleep apnea in systemic hypertension, cardiac arrhythmias, myocardial infarctions, and sudden cardiac death [19].

In a population-based sample of working men and women 30 to 60 years of age, Young et al. [32] show that 4 percent of women and 9 percent of men in this group have 15 or more episodes of sleep apnea per hour of sleep. In 1988, Congress established the National Commission on Sleep Disorders Research "to develop a long-range plan for the use and organization of national resources to deal effectively with sleep disorders research and medicine" [24]. A time-series analysis of sleep apnea variables provides a model for predicting the onset of sleep apnea before it occurs. The analysis is also valuable because it sheds light on the physiological events prior to the onset and on the relationships between the evolutions of the variables.

### 5.2   MODEL STRUCTURE

The sleep-apnea data set contains three continuous variables (HR, CV, SaO2) and one discrete variable (REM). The CV and SaO2 sensors were known to drift slowly over time, and were occasionally rescaled by a technician; hence, their calibrations were known not to be constant over the entire data set.

To discretize the continuous variables, we used KnowledgeSeeker [11], a clustering program based on automatic interaction detection [21]. KnowledgeSeeker produced 7-valued discretizations for each continuous variable.

We then used DYNEMO to generate probable DNM structures from a subset of the database, using the modified K2 metric and the partial ordering imposed by the temporal nature of the model variables. We then refined the model with knowledge of cardiovascular and respiratory physiology, during the process of model fitting and diagnostic checking. The final DNM is shown in Figure 1a. Note that the model has an irregular structure and that $CV_{t+1}$ depends not only on $CV_t$, but also on $CV_{t-1}$. DYNEMO can generate and forecast DNMs with highly irregular structures and noncontemporaneous dependencies that span multiple time points.

### 5.3   MODEL FORECASTS

We present the results of an experiment using the DNM from Section 5.2. We generated the apnea DNM using the first 27,000 time points from the data set, and then used the model to generate the one-step through ten-step-ahead forecasts for the remain-ing 7,000 time points. Figure 2 depicts one-step ahead forecasts for HR, CV, and SaO2, respectively, for the interval $t = [27001, 27400]$.

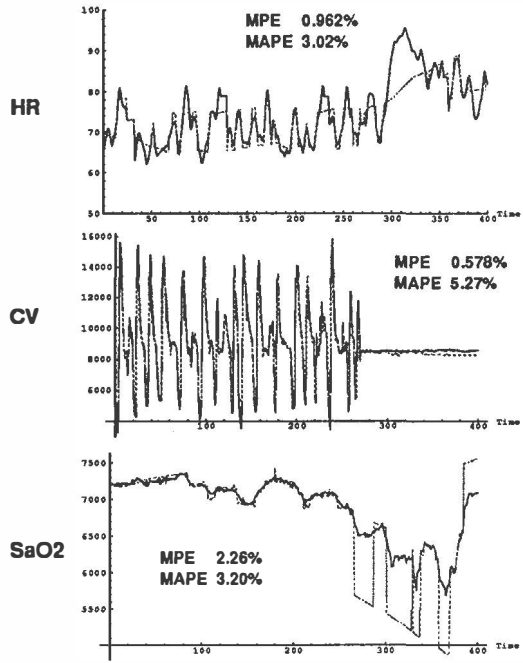

Figure 2: Plots of HR, CV, and SaO2 one-step ahead forecasts over the interval 27001 to 27400. The expected value of the forecast distribution is shown with a dotted line; observed data is depicted by a solid line. Mean prediction errors (MPE) and mean absolute prediction errors (MAPE) are presented for each series.

To analyze our results, we use several measures from statistical forecasting. The *prediction error* at time $t$ is the difference between the observation at time $t$, $O_t$, and the prediction for time $t$, $P_t$. The *percent prediction error (PPE)* at time $t$ is given by $\frac{O_t - P_t}{O_t} \times 100$. The *mean prediction error (MPE)* is given by $\frac{1}{N} \sum_{t=1}^{N} \frac{O_t - P_t}{O_t}$.

The MPE is the sum of the prediction errors normalized by the level of the series measured. Thus, we can compare the MPE between different time series. The MPE measures the *bias* of the model predictions, and therefore, it measures how closely the predictions follow the level of the time series. Statisticians seek values of the MPE that are zero; deviations from zero indicate model bias.

The mean absolute prediction error (MAPE) is given by $\frac{1}{N} \sum_{t=1}^{N} \frac{|O_t - P_t|}{O_t}$. The MAPE measures the *dispersion* of the model predictions. Thus, the MAPE measures how well the predictions capture the turning points in the time series. Whereas the MPE is near zero when the prediction errors cancel, the MAPE may be quite large if the prediction errors are large. Figure 2 gives the MPE and MAPE for HR, CV, and SaO2 for the one-step-ahead predictions in the interval 27001 to 27400.

Notice that in Figure 2, an episode of sleep apnea starts at time 270, corresponding to time point 27270. During the apneic episode, the chest volume is constant, as we expect since the patient has stopped breathing. The blood oxygen saturation, measured by the SaO2, drops during the the apneic period because the patient is not breathing. Also during the apneic period, we observe an increase in the baseline heart rate.

Thus, the three time-series are cross-correlated since concomitant level changes occur during apneic episodes. We also observe concomitant changes during breathing: inspiration (increasing chest volume) causes a decrease in heart rate. This phenomenon is known as *sinus arrhythmia*. Thus, the time series also have cross-correlated cyclicities.

Figure 2 presents the MPE and MAPEs for the one-step ahead forecasts for HR, CV, and SaO2. Because the MPE and MAPEs are dimensionless measures, we can compare their values between different time series. We conclude that there was negligible bias in the HR and CV forecasts. The SaO2 forecasts had a larger bias but the bias is well within the acceptable limit. The MAPE measures the dispersion of the forecasts. The CV forecasts had the largest dispersion, even though of the three forecasted variables, the CV forecasts followed the time-series values closest. The nature of the CV time series explains the wide dispersion of the CV forecasts. The CV time series has rapid, large-amplitude oscillations, which magnify the prediction errors, and therefore, increase the dispersion.

The HR forecasts smooth the real time series: the forecast series truncates peaks and smooths out troughs. However, the forecasts capture turning points in the time series and closely follow the level of the series. The smoothing behavior reflects the forecast's inertia to adapt to the apneic episode, apparent during the time interval between 300 and 350. The CV forecasts follow tightly the real data, capturing all the turning points. The SaO2 forecasts follow the series closely until the apneic episode. At this point, although the forecasts respond to the turning points in the series, the trough levels of the time series were poorly modeled by the forecast series. A plot of the prediction errors (not shown) confirms that trough levels are poorly modeled during the apneic episode: we observe large positive deviations in comparison to the negligible and unbiased prediction errors that precede the apneic episode. The discretization of the time series in the range of the apneic episode, 5700-6500, is too coarse for the forecast values to accurately capture the predictions.

In Figure 3, we show one through ten step-ahead forecasts starting at time 27075. The initial few forecasts follow the real data closely. The prediction error increases as we forecast further into the future.

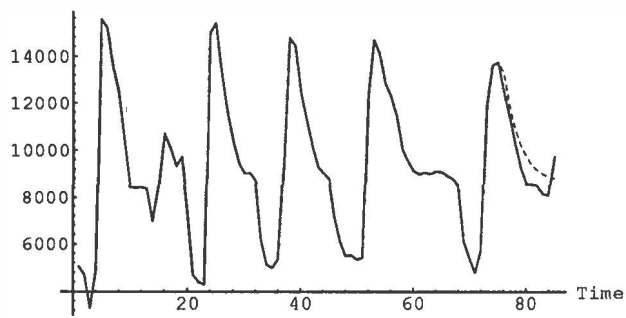

Figure 3: Plot of one-step ahead through 10-step ahead forecasts for CV at $t = 27075$. The solid line is the observed data; the dotted line depicts the expected trajectory of CV for the five second period starting at $t = 27075$, based on **DYNEMO** forecasts.

## 6  CONCLUSIONS

We have been motivated by the need for a time-series model that can be applied to the dynamic domains encountered in complex probabilistic-reasoning applications. DNMs are ideally suited for time-series modeling, forecasting, simulation, and control in domains for which prior knowledge is available about the dynamic forces and relations at work, but is sufficiently complex to preclude a complete specification. These are the domains that we face in probabilistic-reasoning applications.

From the perspective of AI, DNMs inherit the representational virtues of belief networks and influence diagrams. DNMs (1) are explicative models of the domain, as opposed to merely descriptive ones; (2) facilitate the knowledge-engineering tasks of knowledge acquisition and representation; (3) can make normative prescriptions; and (4) are amenable to a variety of efficient inference algorithms. On the other hand, there is too much uncertainty in the time evolution of a dynamic domain for conventional belief-network models to be of value. This uncertainty arises from variations in unmodeled exogenous forces, which are typically too numerous, too difficult, or too obscure to be modeled realistically. If we specify unique conditional-probability distributions to model the dynamic behavior of a complex domain we must sum over all possible states of the unmodeled exogenous influences. Thus, we increase substantially the variance of the conditional probabilities, and therefore, decrease the utility of the model.

From the perspective of time-series analysis, DNMs allow (1) measurement and system equations to be highly nonlinear or specified nonfunctionally; (2) lag influences to be modeled by system equations that are higher-order Markov processes; (3) the specification of disturbance distributions in either functional or table form; and (4) the specification of contemporaneous influences of any nature. Furthermore, probabilistic inference in DNMs is akin to Kalman filtering. Thus, DNMs extend state-space models in every respect. But unlike the problems of tractability and accuracy that we encounter in using Kalman filters with nonnormally distributed disturbances, inference in DNMs is amenable to a number of efficient inference algorithms and to results on the second-order probability distribution of the estimate. By making use of the conditional-independence statements explicit in a belief network, inference algorithms achieve greater efficiency and superior accuracy as compared to all-purpose Monte Carlo algorithms used in nonnormal state-space models.

DNMs extend classical time-series analysis and provide an expressive methodology for ongoing research on temporal probabilistic reasoning. Current research on DNMs includes the convergence analysis of model update methods, the construction of models in the frequency domain, and the development of discrete-event simulation and temporal decision models.

## Acknowledgments

We are grateful to Eric Horvitz for sharing with us his insights into the subject of this paper. This work was supported by the National Science Foundation under grant IRI-9108385, by Rockwell International Science Center IR&D funds, and by the Stanford University CAMIS project under grant IP41LM05305 from the National Library of Medicine of the National Institutes of Health.

## References

[1] B. Abraham and J. Ledolter. *Statistical Methods for Forecasting*. Wiley, New York, 1983.

[2] B. Abramson. ARCO1: An application of belief networks to the oil market. In *Proceedings of the Seventh Conference on Uncertainty in Artificial Intelligence*, pages 1–8, Los Angeles, CA, July 1991. Association for Uncertainty in Artificial Intelligence.

[3] F. Bacchus. *Representing and Reasoning with Probabilistic Knowledge: A Logical Approach to Probabilities*. MIT Press, Cambridge, MA, 1990.

[4] C. Berzuini, R. Bellazzi, and S. Quaglini. Temporal reasoning with probabilities. In *Proceedings of the 1989 Workshop on Uncertainty in Artificial Intelligence*, pages 14–21, Windsor, Ontario, July 1989. Association for Uncertainty in Artificial Intelligence.

[5] R. G. Brown. *Smoothing, Forecasting and Prediction*. Prentice-Hall, Englewood Cliffs, NJ, 1963.

[6] G. Cooper and E. Herskovits. A Bayesian method for the induction of probabilistic networks from data. *Machine Learning*, 9:309–347, 1992.

[7] P. Dagum and R.M. Chavez. Approximating probabilistic inference in Bayesian belief networks. *Pattern Analysis and Machine Intelligence*, 15(3):246–255, 1993.

[8] P. Dagum and A. Galper. Additive belief network models. In *Proceedings of the Ninth Conference on Uncertainty in Artificial Intelligence*, Washington, DC, July 1993. Association for Uncertainty in Artificial Intelligence.

[9] P. Dagum, A. Galper, and E. Horvitz. Dynamic network models for forecasting. In *Proceedings of the Eighth Workshop on Uncertainty in Artificial Intelligence*, pages 41–48, Stanford, CA, July 1992. American Association for Artificial Intelligence.

[10] P. Dagum, A. Galper, and E. Horvitz. Dynamic network models for temporal probabilistic reasoning. Technical Report KSL-91-64, Section on Medical Informatics, Stanford University, Stanford, CA, 1992. Under review for *Pattern Analysis and Machine Intelligence*.

[11] Barry de Ville. Applying statistical knowledge to database analysis and knowledge base construction. KnowledgeWorks Research Systems Ltd., Ottawa, Canada, 1990.

[12] T. Dean and K. Kanazawa. Probabilistic causal reasoning. In *Proceedings of the Fourth Workshop on Uncertainty in Artificial Intelligence*, pages 73–80, Minneapolis, MN, August 1988. American Association for Artificial Intelligence.

[13] T. Dean and K. Kanazawa. A model for reasoning about persistence and causation. *Computational Intelligence*, 5:142–150, 1989.

[14] T. L. Dean and M. P. Wellman. *Planning and Control*. Morgan Kaufmann Publishers, San Mateo, CA, 1991.

[15] J. C. Spall (Ed.). *Bayesian Analysis of Time Series and Dynamic Models*. Marcel Dekker, New York, 1988.

[16] J. Y. Halpern. An analysis of first-order logics of probability. In *Proceedings of the Eleventh International Joint Conference on Artificial Intelligence*, pages 1375–1381, 1989.

[17] A.C. Harvey. *Forecasting, Structural Time Series Models, and the Kalman Filter*. Cambridge University Press, New York, 1990.

[18] E. Herskovits. *Computer-based Construction of Probabilistic Networks*. PhD thesis, Program in Medical Information Sciences, Stanford University, Stanford, CA, 1991.

[19] J. W. Shepard Jr. Hypertension, cardiac arrhythmias, myocardial infarction, and stroke in relation to obstructive sleep apnea. *Clinics in Chest Medicine*, 13:437–458, 1992.

[20] K. Kanazawa. A logic and time nets for probabilistic inference. In *Proceedings of the Ninth National Conference on Artificial Intelligence*, pages 360–365. American Association for Artificial Intelligence, July 1991.

[21] G. V. Kass. An exploratory technique for investigating large quantities of categorical data. *Applied Statistics*, 29:119–127, 1980.

[22] C.R. Kenley. Influence diagram models with continuous variables. Technical Report LMSC-DO67192, Astronautics Division, Lockheed Missiles & Space Company, Sunnyvale, CA, June 1986.

[23] U. Kjaerulff. A computational scheme for reasoning in dynamic probabilistic networks. In *Proceedings of the Eighth Conference on Uncertainty in Artificial Intelligence*, Stanford, CA, 1992. Association for Uncertainty in Artificial Intelligence.

[24] National Commission on Sleep Disorders Research. Wake up America: a national sleep alert. *Government Printing Office*, 1993.

[25] Wolfram Research. *Mathematica 2.0*. Addison-Wesley, Redwood City, CA, 1991.

[26] Y. Shoham. *Reasoning About Change: Time and Causation from the Standpoint of Artificial Intelligence*. MIT Press, Cambridge, MA, 1988.

[27] Y. Shoham and N. Goyal. Temporal reasoning in artificial intelligence. In *Exploring Artificial Intelligence*. Morgan Kaufmann, San Mateo, CA, 1988.

[28] J. Tatman. *Decision Processes in Influence Diagrams: Formulation and Analysis*. PhD thesis, Department of Engineering-Economic Systems, Stanford University, Stanford, CA, 1985.

[29] J. Tatman and R. Shachter. Dynamic programming and influence diagrams. *IEEE Transactions on Systems, Man, and Cybernetices*, 20:365–379, 1990.

[30] A. Weigend and N. Gershenfeld. Time-series competition. In M. Casdagli and S. Eubank, editors, *Nonlinear Modeling and Forecasting: Proceedings Volume XII, Santa Fe Institute Studies in the Sciences of Complexity*, page 17. Santa Fe Institute, Addison-Wesley, September 1990.

[31] M. West and J. Harrison. *Bayesian Forecasting and Dynamic Models*. Springer-Verlag, New York, 1989.

[32] T. Young, M. Palta, J. Dempsey, J. Skatrud, S. Weber, and S. Badr. The occurrence of sleep-disordered breathing among middle-aged adults. *New England Journal of Medicine*, 328:1230–1235, 1993.